%% file: main.tex
\documentclass[10pt,twocolumn,letterpaper]{article}

\usepackage[final]{cvpr}      
\usepackage{booktabs}
\usepackage{pifont}
\usepackage{multirow}
\usepackage{amssymb}
\input{preamble}
\definecolor{cvprblue}{rgb}{0.21,0.49,0.74}
\usepackage[pagebackref,breaklinks,colorlinks,allcolors=cvprblue]{hyperref}

\begin{document}

\title{DINOv3 Meets YOLO26 for Weed Detection in Vegetable Crops}

\author{Boyang Deng and Yuzhen Lu* (luyuzhen@msu.edu)\\
Department of Biosystems \& Agricultural Engineering, Michigan State University\\
East Lansing, MI 48824\\
}
\graphicspath{{./fig}}






\graphicspath{{./fig}}

\maketitle
\input{sec/0_abstract}    
\input{sec/1_intro}
\input{sec/2_method}

\input{sec/3_result}
\input{sec/4_discussion}
\input{sec/5_conclusion}
{
    \small
    \bibliographystyle{ieeenat_fullname}
    \bibliography{reference}
}


\end{document}

%% file: sec/0_abstract.tex
\begin{abstract}
Developing robust models for precision vegetable weeding is currently constrained by the scarcity of large-scale, annotated weed-crop datasets. To address this limitation, this study proposes a foundational crop-weed detection model by integrating heterogeneous datasets and leveraging self-supervised learning. A total of 618,642 crop-weed images were initially collected and subsequently refined to 199,388 filtered images for fine-tuning a DINOv3 vision transformer (ViT-small) through a sequential curation strategy. The fine-tuned DINOv3 backbone was then integrated into YOLO26, serving either as a primary backbone or part of a dual-backbone architecture. A feature alignment loss was introduced in the dual backbone framework to enhance feature fusion with minimal computational overhead. Experimental results show that the proposed DINOv3-finetuned ViT-small-based YOLO26-large achieved up to a +5.4\% $\text{mAP}_{50}$ gain on in-domain images collected in the 2025 season. Moreover, it demonstrated strong cross-domain generalization with $\text{mAP}_{50}$ improvements of +14.0\% on the 2021–2023 season dataset and +11.9\% on the 2024 season dataset, compared to the standard YOLO26-large. Although the DINOv3-YOLO26-large model has 45.6\% more parameters and a $2.9 \times$ increase in inference latency, it maintains real-time performance at $\sim$28.5 frames per second (fps). The curated dataset and software programs developed in this study will be made publicly available.
\end{abstract}

%% file: sec/1_intro.tex
\section{Introduction}
Weeds represent a persistent biological risk to global crop production, and weed management remains a major challenge for yields, food security, and economic sustainability \cite{benjamin2024cereal,huang2025common}. As the most damaging biotic stressor, weeds cause an estimated 34\% yield loss in major crops \cite{kumar2022long} and over 70\% yield loss in vegetable crops \cite{chacko2021integrated}, resulting in approximately \$32 billion in annual global economic losses \cite{kubiak2022problem}. Interactions between weeds and other biotic and abiotic stressors, such as insect pests and nutrient deficiency, can intensify crop yield and quality losses \cite{norris2000interactions, kubiak2022problem}. Conventional broadcast herbicide use causes environmental contamination \cite{fishkis2024ecological} and is increasingly limited by widespread herbicide resistance, which has been documented in 273 weed species worldwide \cite{weedscienceorg}. Weed management is particularly challenging for vegetables, which have limited competitiveness with weeds and herbicide options for weed control \cite{Boyd2022}. These challenges, combined with slow bioherbicide development \cite{duke2024new}, continued dependence on short-residual pre-emergence (PRE) programs \cite{jones2025overlapping}, and unsustainable manual labor \cite{becerra2025weed}, often result in inadequate weed control \cite{zimdahl2024fundamentals} and highlight the need for precision, chemical-reduced weed management strategies.

There is growing interest in implementing precision weed control that integrates machine vision, artificial intelligence (AI), and automation/robotics into weed management \cite{Brainard2023, wang2025precision}. Machine-vision-guided spray systems have been reported to achieve up to 90\% precision with substantially reduced chemical inputs \cite{vijayakumar2023smart}. Other precision weeding strategies are being explored to mitigate chemical dependency. Intelligent mechanical implements such as spring tines and rotary hoes serve as effective complements to chemical control \cite{xiang2024crop}, while laser weeding offers highly localized, plant-level treatments via focused light beams \cite{mutuyimana2025laser}. However, automated or robotic weeders equipped with these actuators rely on accurate, reliable crop-weed detection to function correctly. Inaccurate plant localization or classification can result in unintended crop damage or a failure to suppress weed competition, thereby diminishing the benefits of precision weeding systems.

For machine vision-based weed management, traditional image-processing approaches that rely on geometric, morphological, and color features \cite{thompson1991potential, slaughter2008autonomous} often fail under complex field conditions. The introduction of AI and deep learning has significantly enhanced system reliability and enabled accurate identification of multiple crop and weed types, with YOLO-based one-stage object detectors becoming popular for their efficiency and strong detection performance \cite{vijayakumar2024yolo}. In crop–weed detection, YOLO models consistently outperform traditional methods. \cite{osorio2020deep} reported that YOLOv3 improved lettuce detection F1 scores from 0.79 to 0.89 over an SVM-based pipeline. \cite{wang2022weed25} found YOLOv3, YOLOv5, and Faster R-CNN achieved similar $\text{mAP}_{50}$ (approximately 92\%), with YOLOv5 reaching the highest F1 (0.89). \cite{dang2023yoloweeds} further evalauted 18 YOLO variants (v3--v7) with $\text{mAP}_{50}$ of 88--95\% on a 12-class weed dataset. Using YOLOv10, \cite{deng2025field} reported 80.0\% mAP@50 for crop-weed detection and 84.5\% real-time weed hit rate in in-field lettuce spraying. However, generalization across diverse environments remains difficult. For instance, \cite{deng2024weed} showed YOLOv8, YOLOX, and DINO performed well within the same season ($\text{mAP}_{50}$ 95.7--98.7\%), but cross-season accuracy dropped to 81.5\% for YOLOv8. 

The rapid evolution of advanced AI models is creating new opportunities to enhance plant perception in precision weeding systems. Released in January 2026, YOLO26 introduces non-maximum suppression (NMS)-free inference and removes Distribution Focal Loss, significantly reducing latency on edge devices \cite{sapkota2025ultralytics}. Its ProgLoss and Small-Target-Aware Label Assignment (STAL) mechanisms further improve the detection of small or partially occluded weeds in dense crop rows. Meanwhile, the emerging self-supervised framework DINOv3 employs a high-capacity ViT teacher model trained on 1.7 billion curated images, and with Gram anchoring loss to address feature collapse in dense prediction tasks \cite{simeoni2025dinov3}, it achieves superior performance across diverse computer vision benchmarks even without further fine-tuning. In parallel, transformer-based real-time detectors have demonstrated competitive performance compared to the YOLO family. Very recently, RT-DETRv4 \cite{liao2025rt} addresses the limited semantic representation of lightweight detectors by introducing a Deep Semantic Injector (DSI) to transfer high-level semantic knowledge from a frozen DINOv3 teacher to a real-time CNN student during training. Similarly, DEIMv2 \cite{huang2025real} integrates DINOv3 as the backbone within a DETR framework, and employs a Spatial Tuning Adapter (STA) to convert single-scale ViT outputs into multi-scale feature representations, effectively resolving the resolution mismatch in small object detection. These advances highlight the progress of real-time detectors and the strong visual representation capability of the DINOv3-pretrained ViT model, offering potential to enhance detection accuracy and robustness. Therefore, this study proposes to combine YOLO26 with a DINOv3-pretrained ViT backbone to improve weed detection in precision agriculture scenarios. 

The goal of this study is to develop a robust crop–weed detection model by integrating the strong visual representative power of DINOv3 with the lightweight, high-efficiency YOLO26 detector. Building on carefully curated multi-source datasets, we propose a DINOv3-YOLO26 framework that leverages the rich semantic features learned by DINOv3 while preserving the computational efficiency and real-time suitability of YOLO26. Two architectural configurations are explored: 1) a single-backbone configuration where a DINOv3-pretrained ViT replaces the standard YOLO backbone to directly transfer foundation-level representations into the detection pipeline, and 2) a dual-branch configuration, in which a DINOv3-pretrained ViT serves as an auxiliary backbone alongside the native YOLO backbone and features extracted from both branches are fused at multiple stages to combine global semantic awareness (from DINOv3) with lightweight spatially efficient representations (from YOLO). To ensure effective interaction between the two branches, a feature alignment loss is introduced to harmonize YOLO-based and DINO-based features during joint optimization. The proposed models were evaluated on datasets across multiple growth seasons to assess robustness in real-world field variability. 

By adapting foundation-model-enhanced architectures to crop-weed imagery, this study aims to: 1) reduce reliance on large volumes of manually annotated data, 2) improve cross-season and cross-condition generalization, and 3) maintain lightweight inference suitable for practical field deployment. Accordingly, the specific objectives of this study were as follows to:
\begin{itemize}
    \item Propose a structured data curation pipeline for integrating crop and weed imagery from multiple public sources to fine-tune DINOv3 ViT-small.
    
    \item Propose the DINOv3-YOLO26 framework by integrating the DINOv3-finetuned ViT-small into the YOLO26-large, including single- and dual-branch architectures, with multi-stage feature fusion and a feature alignment loss.
    
    \item Evaluate model performance and robustness by assessing the DINOv3-finetuned ViT-small on plan classification, the proposed DINOv3-YOLO26 architectures on crop-weed detection tasks, and generalization capability across multi-season crop-weed datasets.
\end{itemize}

%% file: sec/2_method.tex
\section{Materials and Methods}

\subsection{Dataset}

Three weed datasets were collected from diverse vegetable fields over multiple seasons from 2021 to 2025 in previous research \cite{deng2025weed, deng2025semi, deng2026improvements, deng2025field}, including the 3SeasonWeedDet10, Weed\&CropDet2024, and Weed\&LettuceDet2025 datasets. The majority of the 2021--2023 data constitute the 3SeasonWeedDet10 dataset \cite{deng2025weed}, which contains 8,436 images and 27,963 weed instances across 10 weed classes. In addition, 2,539 unlabeled images, featuring ambiguous or unknown species without annotations, are included. 
The Weed\&CropDet2024 dataset \cite{deng2026improvements} was collected at the Michigan State University Horticulture Teaching and Research Center (Hort, MI), and the Weed\&LettuceDet2025 dataset was collected both at the same site and at the University of Arizona Yuma Agricultural Center (Yuma, AZ). These datasets include both weeds and vegetable crops such as lettuce, radish, and beets. The 2024 dataset comprises 8,664 images, of which 1,002 are labeled, containing 37,010 annotated instances. The 2025 dataset consists of 28,382 images, with 4216 labeled images containing 37,010 instances. Additional details are provided in \cite{deng2026improvements, deng2025field}. 

\begin{figure*}
  \centering
    \includegraphics[width=0.9\linewidth]{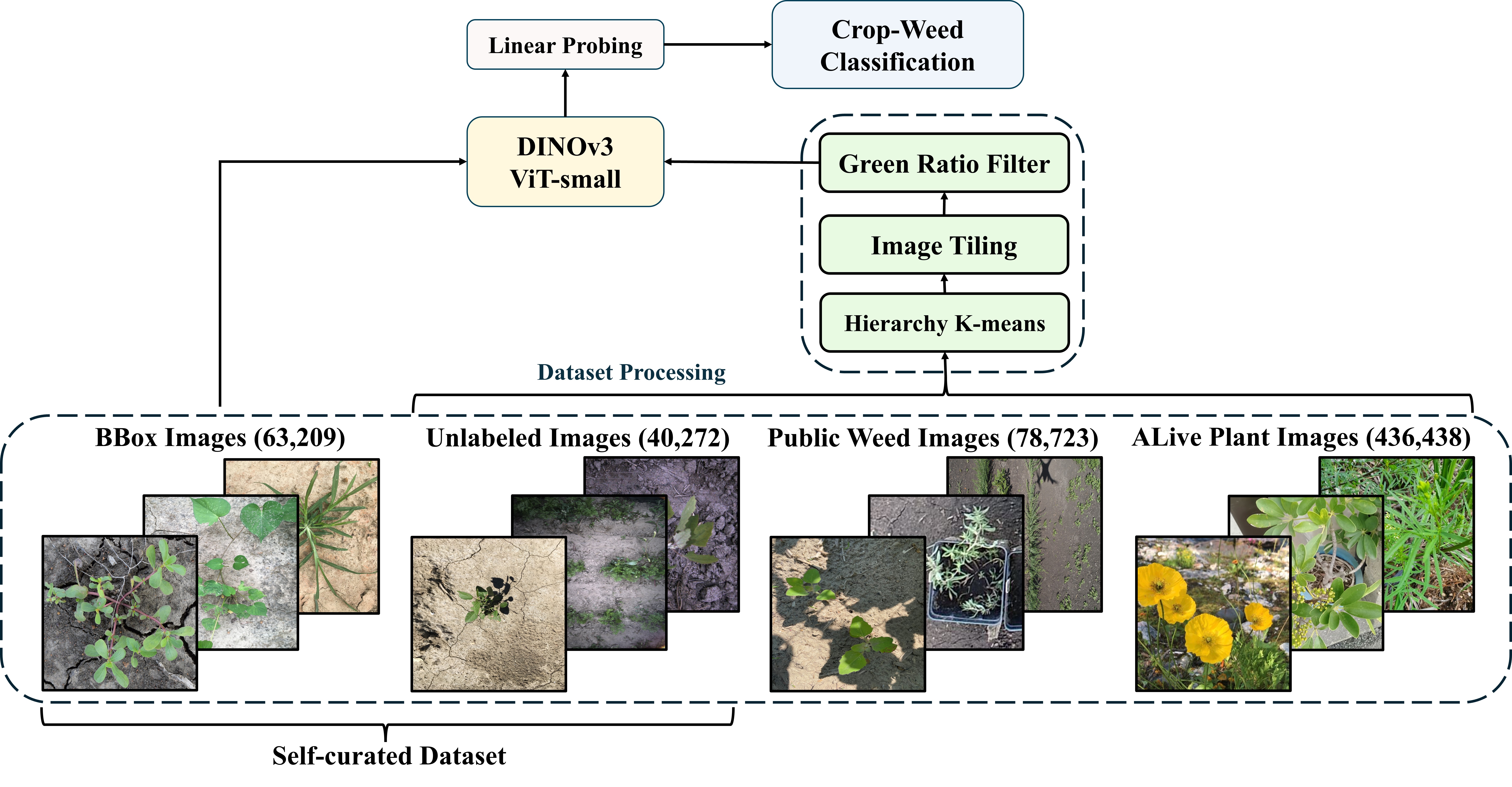}
  \caption{The pipeline of DINOv3 training and evaluation.}
  \label{fig:figure1}
\end{figure*}

With the continuous refinement of data curation strategies aiming at improving robotic weeding vision systems, our image acquisition process evolved substantially over time. The 2021--2023 dataset primarily consists of weed imagery acquired using hand-held smartphones, whereas the 2024 dataset was captured by a mobile platform from the perspective of a typical robotic weeder. However, due to a camera elevation exceeding 50 cm to cover a 75 × 85 cm working area, along with suboptimal lighting conditions, the 2024 dataset images exhibit noticeable motion blur and reduced illumination. Further improvements were implemented in 2025, with a refined camera configuration \cite{deng2025field} positioned approximately 23.5 cm above the ground, resulting in clearer close-range plant imagery. The 2025 lettuce-field images were carefully reviewed, corrected, and fully annotated to include all visible weeds and crops. The resulting refined dataset, Weed\&LettuceDet2025, contains 2,296 images with 11,408 annotated instances, including 3,372 lettuces and 8,036 weeds. 

\begin{figure*}
  \centering
    \includegraphics[width=0.85\linewidth]{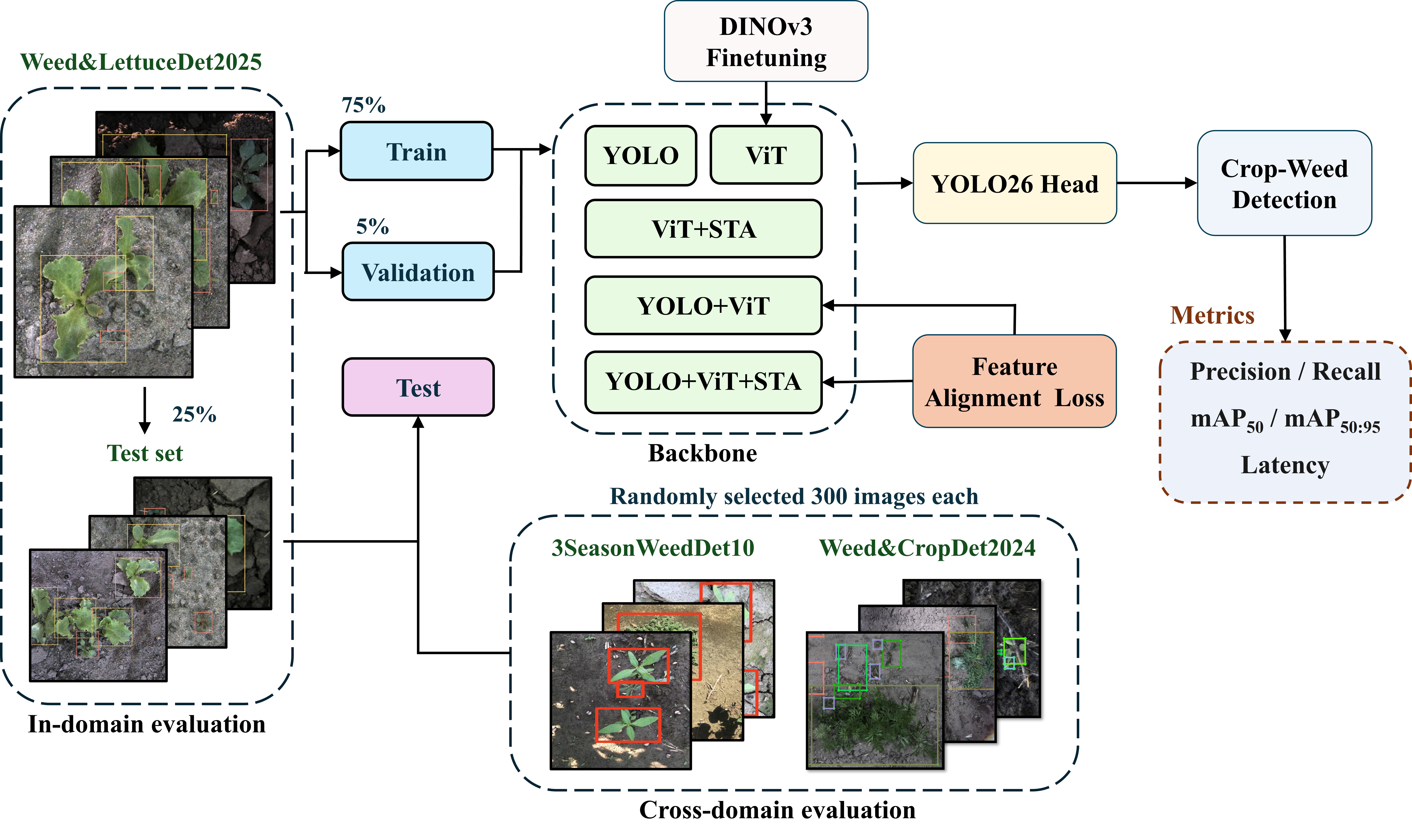}
  \caption{The training pipeline of the DINO-YOLO framework for weed detection.}
 \label{fig:figure2}
\end{figure*}

For cross-domain weed detection performance assessment (Figure \ref{fig:figure2}), 300 images were randomly selected from each of the 3SeasonWeedDet10 (2021--2023) and Weed\&CropDet2024 datasets. These images were re-examined and relabeled, resulting in 2,463 and 10,843 annotated instances, respectively. The 2024 dataset typically exhibits a higher instance density per image due to the larger working. The refined 2025 lettuce images were subsequently used to train the DINO-YOLO26 detection model, followed by in-domain evaluation on the 2025 dataset and cross-domain evaluation on the 2021--2024 datasets.

To further enhance DINOv3 fine-tuning, additional public and self-curated datasets were incorporated. First, 78,723 RGB images from 36 crop-weed datasets reviewed in \cite{deng2024weed} were aggregated into a unified public collection. Second, 436,438 crop images were obtained from ALive \cite{nawaz2025agriclip}, which integrates 14 agricultural crop classification datasets. Third, our 2021--2025 dataset contributed 40,272 unlabeled crop-weed images, along with 63,209 bounding-box-cropped images extracted from all labeled images (excluding those reserved for weed detection). In total, 618,642 images were curated from these three sources for DINOv3 finetuning and will be further processed according to the data curation pipeline described in the next section. Notably, fine-tuning DINOv3 using bounding-box-cropped images rather than direct weed detection improves tolerance to annotation noise, facilitating the future integration of external labeled datasets from other sources with minimal additional verification effort. Figure \ref{fig:figure1} illustrates the DINOv3 fine-tuning pipeline, including data curation, model training, and linear probing-based evaluation \cite{haochen2021provable} for crop-weed classification.

\subsection{DINOv3 Fine-Tuning}
DINOv3 trains a ViT-7B model using 1.7 billion unlabeled images and introduces a novel Gram Anchoring loss to mitigate the feature collapse issue within dense prediction tasks, including detection, segmentation, and depth estimation \cite{simeoni2025dinov3}. Through its three training phases, i.e., pretraining, Gram anchoring, and high-resolution adaptation, it can learn robust and transferable visual representations. These characteristics make it well-suited for data-efficient weed detection, enabling effective adaptation with limited annotation. Moreover, its generalized representations provide robustness across varying plant species and field conditions.

As demonstrated by \cite{oquab2023dinov2, simeoni2025dinov3}, systematic data curation often yields superior performance compared to direct utilization of raw data. Accordingly, a standardized data curation pipeline was implemented to integrate imagery from diverse sources for DINOv3 fine-tuning. First, all previously curated bounding-box images were directly incorporated into the training process. These bounding boxes that are centered on a single plant naturally align with the DINO objective of matching local and global representations, thereby facilitating learning. Inspired by \cite{oquab2023dinov2}, the public crop-weed collection and ALive plant images were independently refined using a hierarchical K-means-based pipeline \cite{vo2024automatic}. The three-level pyramid consisted of 512, 2,048, and 8,192 clusters, balancing performance and computational efficiency. Subsequently, one-tenth of the clustered images were sampled from the pyramid to reduce the scale of the unlabeled data. These selected images were then tiled into $518 \times 518$ patches with 20\% overlap. Background patches were filtered based on vegetation (green) ratio requiring at least 20\% green pixel coverage, following \cite{benito2025vision}. This refinement process produced 47,204 images from the public weed collection, 59,426 from ALive, and 29,499 from the self-curated unlabeled weed dataset, totaling 136,129 images. Combining with 63,209 bounding-box images, the final dataset for DINOv3 fine-tuning comprised 199,338 images.

For fine-tuning, the ViT-small model was selected due to its lightweight design as the smallest ViT model released in \cite{simeoni2025dinov3}, as well as to accommodate the limited computational resources in this study. After training, linear probing was done to evaluate the ViT-small backbone on plant classification using multi-species bounding-box images from the 2021--2025 dataset. The combined dataset included 8,427 images across 23 classes. In parallel with the subsequent detection experiment, this experiment directly evaluates the quality of the learned global visual representations.

\subsection{DINO-YOLO26 Framework}

YOLO26 \cite{jocher2026ultralytics} represents the latest evolution in the YOLO family, designed to streamline real-time detection on edge devices such as weeding robots. By removing Distribution Focal Loss (DFL) and introducing native NMS-free inference, it significantly cuts latency and simplifies deployment \cite{sapkota2025ultralytics}. The MuSGD optimizer is employed to enhance training stability and accelerate convergence. Furthermore, YOLO26 incorporates STAL and Progressive Loss Balancing (ProgLoss) to improve small object detection. In this study, the large variant of YOLO26 was selected because its model size is comparable to ViT-small, providing a favorable trade-off between detection speed and accuracy \cite{huang2025real}.

DEYOLO \cite{chen2024deyolo} employs a dual-branch backbone to extract RGB and infrared image features separately, using semantic and spatial enhancement modules to fuse modalities, dynamically recalibrating channels and sharpening spatial features to emphasize salient information. Inspired by DEYOLO \cite{chen2024deyolo} and the approach in \cite{huang2025real}, which integrates a DINOv3-trained ViT-small into a DETR structure using STA to convert single-scale ViT outputs into multi-scale features, this study adopts the ViT-small as the backbone of the YOLO26 detector.

Specifically, a similar dual-branch backbone architecture is proposed to integrate DINOv3 ViT features with YOLO representations. In one variant, ViT with STA serves as the integrated backbone for YOLO26, maintaining the original ViT output feature dimensions of 384. Subsequently, $1 \times 1$ convolutional layers are appended after the ViT backbone to produce feature dimensions compatible with the YOLO detection head (e.g., 512 channels for the YOLO-large version). Additionally, a dual-branch YOLO26 backbone is designed to combine the original YOLO26 backbone with the ViT-small. To enable multi-scale information fusion, features from YOLO backbone stages P3, P4, and P5 are fused with ViT layers 5, 8, and 11, respectively, consistent with the layer selection strategy in \cite{huang2025real}. To ensure effective interaction between the two branches, a feature alignment loss is introduced to harmonize YOLO-based and ViT-based features, defined as follows:

\begin{equation}
    \mathcal{L}_{\text{Align}} = \sum MSE (\mathbf{X}_\textit{YOLO}^{(i)} - \mathbf{X}_\textit{ViT}^{(j)} )
\end{equation}

where $i$ is the output layer index of the YOLO backbone (P3, P4, P5), $j$ is the corresponding ViT layer index (5, 8, 11). This loss is added to the original YOLO loss with an empirical weighting factor of 1.0 without dominating the detection loss.

\subsection{Experiments}
This study was implemented using PyTorch (v2.5.1), Torchvision (v0.20.1), CUDA (v12.1). All experiments were conducted on an NVIDIA RTX A6000 GPU with 48 GB memory. Figure \ref{fig:figure2} presents the overall training and evaluation pipeline used to compare the DINO-YOLO26 variants against the standard YOLO26 baseline.

\subsubsection{Linear Probing Evaluation}
Preliminary experiments showed that DINOv3-trained ViTs achieved over 96\% average accuracy for both official and fine-tuned ViT models using a 7:3 train-test split on binary lettuce-weed classification with the 2025 lettuce bounding-box images, with differences below 0.5\%. Even under a few-shot scenario (1:9 split), accuracy remained around 94\%, again with minimal differences. To better differentiate the capability of the fine-tuned model, all original crop-weed classes (excluding the generic weed class) were used, forming a 23-class dataset comprising 8427 bounding-box images. A 7:3 split was used for training and testing, and linear probing was conducted for 30 epochs with a batch size of 128. The learning rate was set to 0.05, while other hyperparameters followed the default DINOv3 linear probing configuration.

\subsubsection{DINOv3 Fine-Tuning}
The ViT-small model was trained using DINOv3 on 199,338 curated weed images, substantially fewer than the 1.7 billion images used for the ViT-7B model, and with significantly fewer parameters. To match the scale of ViT-small, both the DINO and iBOT heads employed 4,096 prototypes, a bottleneck dimension of 384, a hidden dimension of 2,048, and the drop path rate of the student model set to 0.1. During pretraining, a batch size of 144 was used for 50,000 steps, then reduced to 96 for 10,000 Gram anchoring steps, and further reduced to 4 for 5,000 High-Resolution Adaptation steps. The reduced batch size was necessary due to the high memory demand of multiple high-resolution cropping operations. All other hyperparameters followed the official DINOv3 recipe.

\subsubsection{YOLO26 Training Strategy}
For detection, models were trained on the 2025 lettuce dataset using a 75:5:25 split for training, validation, and testing, respectively. In training YOLO26, ProgLoss unexpectedly underperformed compared to the DFL box regression loss on our dataset. Additionally, removing the attention mechanism from the final C3k2 block improved detection accuracy. Consequently, a modified YOLO26 head was proposed as a stronger baseline, and subsequent DINO-YOLO26 integrations were built on this modified architecture. Although MuSGB is recommended in YOLO2026, standard SGD yielded slightly better performance on the lettuce dataset. Moreover, because the ViT backbone does not align with the assumptions of the Muon optimizer \cite{liu2025muon}, SGD was employed for all subsequent training.

\subsubsection{DINO--YOLO26 Integration}
The DINO-YOLO26 models were evaluated using both official and fine-tuned ViT-small weights. Empirical results showed that only initializing YOLO26 heads with pre-trained COCO weights led to inferior performance compared to training from scratch on the weed dataset; consequently, all experiments were trained from scratch. Furthermore, unfreezing the ViT backbone yielded superior results compared to freezing it; hence, the ViT backbone parameters remained trainable throughout detection training. The input resolution was increased to $800 \times 800$ from the default $640 \times 640$ to better accommodate high-resolution (900--1400) imagery while maintaining computational efficiency. Given that the architectural concept and STA module were adapted from \cite{huang2025real}, a consistent training duration of 68 epochs was employed. To ensure compatibility with DINOv3's feature space, Z-norm preprocessing was applied in the modified architecture, instead of the standard pixel scaling used in the baseline YOLO26. All other hyperparameters adhered to the default YOLO26 configuration.

\subsubsection{Cross-Domain Evaluation}
To assess the cross-domain generalization, additional evaluations were conducted using 300 randomly sampled images from the 2021--2023 season dataset (3SeasonWeedDet10) \cite{deng2025weed}) and 300 images from the 2024 dataset (Weed\&CropDet2024) \cite{deng2026improvements}). Since the 3SeasonWeedDet10 dataset contains only weed images and the crops in the Weed\&CropDet2024 dataset (beets, radishes, and turnips) differ from the lettuce in the 2025 season data, single-class "Weed" detection was performed for both datasets to ensure consistency. 

\subsection{Metrics}
For the binary crop-weed classification task in linear probing, Top-1 accuracy was used to evaluate the DINOv3-trained ViT-small model before and after fine-tuning on the curated weed dataset, calculated as follows:

\begin{equation}
    \text{Accuracy (\%)} = \frac{\#TP + \#TN}{\#TP + \#TN + \#FP + \#FN}
\end{equation}
Where $\#TP,\, \#FP,\, \#TN, \text{and}\, \#FN$ denote the numbers of true positives, false positives, true negatives, and false negatives, respectively, based on the predicted and ground-truth labels for each bounding-box image.

Plant detection performance was evaluated using standard object detection metrics, including precision, recall,  and mAPs ($\text{mAP}_{50}$, and $\text{mAP}_{50:95}$). Precision is the ratio of correctly positive predictions to the total predicted positives, and recall is the ratio of correctly positive predictions to the total number of true positives. The $\text{mAP}_{50}$ was computed as the mean average precision across the two classes of crop and weed at an IoU threshold of 0.5, while $\text{mAP}_{50:95}$ was computed across a range of IoU thresholds from 0.5 to 0.95 in steps of 0.05. These metrics are calculated below:

\begin{equation}
  \text{Precision (\%)} = \frac{\#TP}{\#TP + \#FP} \times 100\%
  \label{eq:precision}
\end{equation}

\begin{equation}
  \text{Recall (\%)} = \frac{\#TP}{\#TP + \#FN} \times 100\%
  \label{eq:recall}
\end{equation}

\begin{equation}
  \text{mAP (\%)} = \frac{1}{N} \sum_{i=1}^{N} \text{AP}_i \times 100\%
  \label{eq:mAP}
\end{equation}

Where $\#TP,\, \#FP,\, \#TN, \text{and}\, \#FN$ have the same definitions as before except used for detection. A detection was counted as correct if it matched the ground-truth label, exceeded a confidence threshold of 0.25, and had an intersection-over-union (IoU) of at least 0.5. The number of classes, N, was set to two: one crop class (lettuce) and a generic “weed” class that included all weed species.

%% file: sec/3_result.tex
\section{Results}

\subsection{Plant Classification by DINOv3}

Table \ref{tab:table1} shows the linear probing accuracy of the DINOv3-trained ViT-small on multi-species plant classification. While the original ViT provides a robust feature extraction foundation with an average accuracy of 87.67\%, fine-tuning the ViT on weed imagery improves the performance to 89.94\% after gram anchoring, surpassing the baseline by +2.27\%. Crop classes typically achieve $\sim$10\% higher accuracy than weeds, highlighting the greater variability and challenge associated with weed recognition. Interestingly, after high-resolution adaptation, the accuracy drops to 82.50\%, suggesting a reduction in the global visual representation capability. This decrease may be due to the limited batch size of 4 in this training phase, constrained by the 48 GB GPU memory used in this study. However, subsequently, ViT weights from each training phase were evaluated as the DINO-YOLO backbone for weed detection. The high-resolution trained ViT achieved the best performance and was therefore adopted in following DINO*-YOLO26*.

\begin{table*}[ht]
    \centering
    \caption{Linear probing accuracy of the DINOv3-trained ViT-Small backbone. The finetuning is based on the DINOv3 -trained official ViT-small model, consisting of three consecutive phases:  pretraining, gram anchoring, and high-res adaptation. Crop refers to the average accuracy across all crop classes, while weed refers to the mean accuracy for all weeds.}
    \label{tab:table1}
    \setlength{\tabcolsep}{12pt} 
    \renewcommand{\arraystretch}{1.2}

    \begin{tabular}{lccc}
    \toprule
    \textbf{Backbone Configuration} & \textbf{Crop (\%)} & \textbf{Weed (\%)} & \textbf{Average (\%)} \\
    \cmidrule{1-4}

    Official ViT-Small     & 92.2 & 83.1 & 87.7 \\
    + Pretraining          & 95.0 & \textbf{83.7} & 89.3 \\
    + Gram Anchoring       & \textbf{96.6} & 83.3 & \textbf{89.9} \\
    + High-Res Adaptation  & 88.6 & 76.4 & 82.5 \\
    \bottomrule
    \end{tabular}
\end{table*}

Table \ref{tab:table2} summarizes the results of the ablation study on Weed\&LettuceDet2025. The default DINOv3-trained ViT-small was utilized to focus on model modification contributions. The original YOLO26 model (backbone and head) received the lower bound of performance with 84.9\% $\text{mAP}_{50}$ and 63.2\% $\text{mAP}_{50:95}$. The modified YOLO heads trained with DFL loss achieved improvements of +3.4\% $\text{mAP}_{50}$ and +3.3\% $\text{mAP}_{50:95}$, providing empirical evidence that DFL is advantageous for weed datasets and that removing the attention block of the last C3k2 module enhances model accuracy. The ViT-based backbone outperformed the YOLO26 backbone, attaining the highest precision of 89.7\% and $\text{mAP}_{50}$ of 91.9\%, and with the further addition of STA, it received the highest $\text{mAP}_{50:95}$ of 72.6\%, which is a +9.4\% gain in $\text{mAP}_{50:95}$ compared to the original YOLO26 model. For the dual-branch backbone models, although assignment loss can improve results to match the single ViT-backbone results, the single backbone DINO-ViTs outperform their dual-branch backbone versions in model complexity and latency, which may indicate that the feature fusion for dual backbones needs further exploration. 

\begin{table*}[h]
    \centering
    \caption{Ablation study to evaluate the contributions of different components in the DINO-YOLO framework. Y-H refers to the YOLO26-large head, and Y-H* indicates the modified YOLO26-large head. Y-B denotes the original YOLO26-large backbone. ViT represents the DINOv3-pretrained ViT-small. STA is the auxiliary branch added to the ViT, and Align refers to the feature alignment loss used in the dual-branch backbones. Lat. stands for latency, and Param. denotes model parameters.}
    \label{tab:table2}
    \resizebox{\textwidth}{!}{%
        \begin{tabular}{cccccccccccc}
            \toprule
            \multicolumn{6}{c}{\textbf{Components}} & \multicolumn{6}{c}{\textbf{Metrics}} \\
            \cmidrule(lr){1-6} \cmidrule(lr){7-12}
            \textbf{Y-H} & \textbf{Y-H*} & \textbf{Y-B} & \textbf{ViT} & \textbf{STA} & \textbf{Align} & \textbf{Precision (\%)} & \textbf{Recall (\%)} & \textbf{$\text{mAP}_{50}$ (\%)} & \textbf{$\text{mAP}_{50:95}$ (\%)} & \textbf{Lat. (ms)} & \textbf{Param. (M)} \\
            \midrule
            \checkmark & & \checkmark & & & & 80.8 & 79.7 & 84.9 & 63.2 & 12.0 & 24.8 \\
            & \checkmark & \checkmark & & & & 86.3 & 84.3 & 89.8 & 68.9 & 12.4 & 24.5 \\
            & \checkmark & & \checkmark & & & \textbf{89.7} & 87.2 & \textbf{91.9} & 71.6 & 35.1 & 36.1 \\
            & \checkmark & & \checkmark & \checkmark & & 87.7 & 87.8 & 91.5 & \textbf{72.6} & 36.1 & 36.7 \\
            & \checkmark & \checkmark & \checkmark & & & 89.6 & 86.5 & 91.3 & 71.0 & 41.6 & 50.2 \\
            & \checkmark & \checkmark & \checkmark & & \checkmark & 88.0 & \textbf{87.9} & 91.8 & 71.6 & 41.9 & 50.2 \\
            & \checkmark & \checkmark & \checkmark & \checkmark & & 89.2 & 87.0 & 91.6 & 71.0 & 42.8 & 51.0 \\
            & \checkmark & \checkmark & \checkmark & \checkmark & \checkmark & 89.3 & 87.0 & 91.7 & 71.0 & 42.9 & 51.0 \\
            \bottomrule
        \end{tabular}%
    }
\end{table*}

\subsection{Crop-Weed Detection}
Table \ref{tab:table2} shows incorporating STA into the single ViT backbone variant (DINO-YOLO26*) yielded the highest $\text{mAP}_{50:95}$. To verify the robustness of this improvement, three additional replications were conducted for models with and without STA. As summerized in Table \ref{tab:table3}, the base DINO-YOLO26* achieved 92.3\% $\text{mAP}_{50}$ and 72.3\% $\text{mAP}_{50:95}$, whereas the STA-enhanced version obtained 92.0\% $\text{mAP}_{50}$ and 72.4\% $\text{mAP}_{50:95}$. The negligible performance difference suggests that the STA module offers no statistically significant performance gain while slightly increasing inference latency. Therefore, subsequent experiments adopt the DINO-YOLO26 architecture without the STA module. 

Although DINO-YOLO26* contains only 45.6\% more parameters than the original YOLO26, its inference time is approximately 2.9 times longer (12.0 ms vs. 35.1 ms). This increase reflects the computational intensity of the self-attention-based ViT framework, which remains a key efficiency bottleneck for future optimization. Nevertheless, DINO-YOLO26* maintains real-time performance ($\sim$28.5 fps), comparable to previous YOLO architectures such as YOLOv8 under similar input configurations \cite{deng2025development}.

\begin{table*}[!t]
    \centering
    \caption{Performance comparison of YOLO26 and DINOv3-YOLO26 models across different year datasets. Three replications are used. YOLO26* indicates the modified YOLO26 model trained with DFL loss, and DINO* denotes that the DINO-ViT model is fine-tuned on the weed dataset. Superscripts denote the Compact Letter Display (CLD) generated from Tukey’s HSD multiple comparison test, where groups sharing a letter are not significantly different.}
    \label{tab:table3}
    \setlength{\tabcolsep}{6pt} 
    \renewcommand{\arraystretch}{1.2} 

    \begin{tabular}{llcccc}
        \toprule
        \textbf{Dataset} & \textbf{Model} & \textbf{Precision (\%)} & \textbf{Recall (\%)} & \textbf{$\text{mAP}_{50}$ (\%)} & \textbf{$\text{mAP}_{50:95}$ (\%)} \\
        \midrule
        
        \multirow{3}{*}{2025} 
        & YOLO26 & 81.8 & 81.0 & $86.9^{b}$ & $66.2^{b}$ \\
        & YOLO26* & 86.8 & 84.1 & $90.3^{a}$ & $69.4^{ab}$ \\
        & DINO-YOLO26* & \textbf{89.6} & 87.2 & $\textbf{92.3}^{a}$ & $\textbf{72.3}^{a}$ \\
        & DINO*-YOLO26* &  88.6 & \textbf{87.7} & $92.2^{a}$ & $71.4^{a}$ \\
        \cmidrule{1-6}

        \multirow{3}{*}{2024} 
        & YOLO26 & 42.8 & 30.8 & $29.6^{c}$ & $14.6^{b}$ \\
        & YOLO26* & 44.4 & 34.9 & $33.9^{b}$ & $16.2^{c}$ \\
        & DINO-YOLO26* & 53.6 & 38.8 & $40.5^{a}$ & $19.5^{a}$ \\
        & DINO*-YOLO26* & \textbf{53.7} & \textbf{40.1} & $\textbf{41.5}^{a}$ & $\textbf{19.8}^{a}$ \\
        \cmidrule{1-6}

        \multirow{3}{*}{2021--2023} 
        & YOLO26 & 55.1 & 42.9 & $42.5^{c}$ & $25.7^{b}$ \\
        & YOLO26* & 58.4 & 44.8 & $46.3^{c}$ & $28.2^{b}$ \\
        & DINO-YOLO26* & 63.9 & 49.9 & $51.8^{b}$ & $31.0^{a}$ \\
        & DINO*-YOLO26* & \textbf{65.2} & \textbf{54.4} & $\textbf{56.5}^{a}$ & $\textbf{33.1}^{a}$ \\
        \bottomrule
    \end{tabular}
\end{table*}

Table \ref{tab:table3} further presents the detection performance across weed datasets collected in different seasons, highlighting the impact of domain shift. The in-domain 2025 lettuce dataset yielded the highest accuracy, with YOLO2026 reaching 86.9\% $\text{mAP}_{50}$ and 66.2\% $\text{mAP}_{50:95}$, significantly outperforming cross-domain results. In cross-domain evaluations, the 2021--2023 dataset outperformed the 2024 dataset by more than 10\% across all metrics, which may be attributable to the former consisting of high-resolution smartphone images compared to the latter with blurred, lower-quality images. 

The modified YOLO26 (YOLO26*) surpassed the baseline YOLO26, while DINO-YOLO26* consistently outperformed both across all datasets. Compared to the original YOLO26, DINO-YOLO26* achieved superior performance, with gains of +5.4\% $\text{mAP}_{50}$ and +6.2\% $\text{mAP}_{50:95}$ on the 2021--2023 dataset, +10.9\% $\text{mAP}_{50}$ and +4.9\% $\text{mAP}_{50:95}$ on the 2024 dataset, and +9.3\% $\text{mAP}_{50}$ and +5.3\% $\text{mAP}_{50:95}$ on the 2025 lettuce dataset. Interestingly, although the fine-tuned ViT-small model (DINO*-YOLO26*) yielded marginally lower in-domain performance compared to the official DINOv3-pretrained ViT-small (DINO-YOLO26*), it demonstrated enhanced cross-domain generalization, achieving +4.7\% and +1.0\% $\text{mAP}_{50}$ improvements in the 2021--2023 and 2024 datasets, respectively. This suggests that task-specific fine-tuning may enhance robustness to domain variation. Finally, the statistical analysis of $\text{mAP}_{50}$ and $\text{mAP}_{50:95}$ confirms that DINO-based YOLO26 models significantly outperform the original YOLO26 consistently; however, performance differences within the DINO-YOLO26 or YOLO26 variants are mostly statistically insignificant.

%% file: sec/4_discussion.tex
\section{Discussion}

This study presents a promising hybrid detection framework, DINO-YOLO26, which surpasses standard YOLO26 in crop-weed detection by utilizing a DINOv3-trained ViT model. This improvement aligns with existing literature indicating that DINO-based YOLO models excel in small object detection and data-efficient tasks by leveraging the global receptive field of transformers \cite{p2025dinoyoloselfsupervisedpretrainingdataefficient}. Although the hybrid architecture increases the total model parameters by 45.6\%, it exhibits approximately $2.9 \times$ slower inference speeds. This latency bottleneck is primarily attributed to the self-attention mechanism inherent in the ViT, which demonstrates quadratic computational complexity with respect to token length (image size) \cite{dosovitskiy2020image, khan2022transformers}. Future work could bridge the latency gap between ViT and the original YOLO by implementing model compression strategies, such as token pruning \cite{zhu2021vision} or knowledge distillation as used in the original DINOv3 \cite{simeoni2025dinov3}. Additionally, future work may explore the feasibility of utilizing the default YOLO backbone for self-training methods like DINOv3.

The high-resolution training phase of DINOv3 yielded inferior weed classification results compared to earlier phases, a discrepancy likely attributable to the constrained batch size of 4. The limited computational resources in this study may have hindered the efficacy of DINOv3 training, particularly during high-resolution adaptation, resulting in reduced accuracy. Future research aims to scale computational capacity to approach the original batch size of 4096 as in \cite{huang2025real}. Unlike the broad improvements reported in \cite{simeoni2025dinov3}, DINOv3 fine-tuning of the ViT-small backbone led to stronger cross-domain performance but only marginal in-domain gains, indicating further analysis is needed. Furthermore, the dual-branch backbones achieved accuracy comparable to single ViT backbones, suggesting that further investigation into advanced feature fusion mechanisms is warranted, such as cross-attention feature fusion \cite{carion2020end} or fusion strategies employed in multi-modal detection frameworks utilizing dual-branch architectures \cite{zhao2023cddfuse, chen2024deyolo}.

%% file: sec/5_conclusion.tex
\section{Conclusion}

This study proposes a novel DINO-YOLO26 framework for enhanced crop-weed detection by replacing the original YOLO backbone with a DINOv3-pretrained ViT. Large-scale public datasets were curated, yielding 199,338 images for fine-tuning DINOv3-pretrained ViT-small. The fine-tuned DINOv3 was integrated into YOLO26-large with both single and dual-branch backbone designs, with a feature alignment loss introduced to facilitate feature fusion in the dual-branch architecture. Results show that DINO-YOLO26* improved $\text{mAP}_{50}$ by 5.4\% (86.9 vs. 92.3\%) and $\text{mAP}_{50:95}$ by 5.2\% (66.1 vs. 71.4\%). The fine-tuned ViT-based YOLO26-large (DINO*-YOLO26*) exhibited enhanced cross-domain weed detection generalization, with $\text{mAP}_{50}$ gains of 14.0\% on the 2021--2023 season dataset and 11.9\% on the 2024 dataset, indicating improved robustness to domain variations. Although the proposed model increases parameter count by 45.6\% and inference time by $2.9 \times$, it still maintains real-time efficiency at approximately 28.5 fps. 